\documentclass[journal]{IEEEtran}
\usepackage{iftex}
\usepackage{amsmath}
\ifPDFTeX
    \usepackage{amssymb,amsfonts}
\else
    \usepackage{fontspec}
    \usepackage{unicode-math}
    \defaultfontfeatures{Ligatures=TeX}
    \IfFontExistsTF{Cambridge Math}{
        \setmathfont{Cambridge Math}
    }{
        \IfFontExistsTF{Cambria Math}{
            \setmathfont{Cambria Math}
        }{
            \IfFontExistsTF{STIX Two Math}{
                \setmathfont{STIX Two Math}
            }{
                \setmathfont{latinmodern-math.otf}
            }
        }
    }
\fi
\usepackage{array}
\usepackage[caption=false,font=normalsize,labelfont=sf,textfont=sf]{subfig}
\usepackage{textcomp}
\usepackage{stfloats}
\usepackage{url}
\usepackage{graphicx}
\usepackage{cite}
\usepackage{multirow}
\newcommand{\best}[1]{\textbf{#1}}

\begin{document}

\title{EAGG: Embodiment-Aligned Grasp Generation via Geometry-Aware Graph Conditioning}

\author{Wanhao~Niu, Qiyan~Ke, Yuan~Sun, Hao~Sun, Jie~Xu, Muyuan~Ma, Ruiqi~Hu, and Fuchun~Sun\thanks{Wanhao Niu, Qiyan Ke, Hao Sun, and Fuchun Sun are with the Department of Computer Science and Technology, Tsinghua University, Beijing, China. Yuan Sun, Jie Xu, Muyuan Ma, and Ruiqi Hu are with Beijing Moce Future Technology Co., Ltd., Beijing, China. Corresponding author: Fuchun Sun.}}

\markboth{}{Niu \MakeLowercase{\textit{et al.}}: EAGG: Embodiment-Aligned Grasp Generation}

\maketitle

\begin{abstract}
Cross-end-effector grasp generation seeks a unified model that generalizes across objects and across embodiments ranging from parallel grippers to dexterous end effectors. Existing grasp generators are typically designed for a fixed embodiment or encode embodiment identity with a static descriptor, which weakens transfer when topology, actuation coupling, and contact geometry differ substantially. We present EAGG, an embodiment-aligned grasp generator that represents each embodiment with a topology-aware end-effector graph and an embodiment-specific low-dimensional end-effector control space. A frozen end-effector-cognition backbone converts the current articulated state into geometry-aware tokens that act as a reusable morphology prior, and iterative geometry injection refreshes these tokens throughout sampling so that conditioning remains synchronized with the evolving end-effector geometry. On the MultiGripperGrasp benchmark, EAGG reaches 56.17\% average success across six training end effectors, remaining within 1.10 percentage points of specialized training while preserving transfer to finetuning and zero-shot end effectors. Iterative geometry injection further reduces the pooled median contact distance from 0.239\,cm to 0.189\,cm. These results show that cross-end-effector grasp generation is strengthened by aligning embodiment structure inside a shared generator rather than suppressing embodiment differences. Code is available at \url{https://github.com/wanhaoniu/EAGG}.
\end{abstract}


\section{Introduction}

Cross-end-effector grasp generation seeks a single model that can synthesize grasps for heterogeneous end effectors while preserving generalization across novel objects. The setting is more demanding than fixed-end-effector grasping because the generator must solve two coupled problems at once: it must infer how an object should be grasped and how that grasp should be expressed for a particular embodiment. Fixed-end-effector methods such as Dex-Net \cite{mahler2017dexnet}, PointNetGPD \cite{liang2018pointnetgpd}, 6-DoF GraspNet \cite{mousavian20196}, S4G \cite{qin2020s4g}, and Contact-GraspNet \cite{sundermeyer2021contact} deliver strong performance when the embodiment is fixed, but they do not provide a unified mechanism for transferring across heterogeneous end effectors.

This limitation becomes increasingly restrictive in manipulation systems that swap hardware within the same workflow. DexGraspNet \cite{Wang2023DexGraspNet} and DexGraspNet~2.0 \cite{zhang2024dexgraspnet} expanded dexterous supervision, UniDexGrasp \cite{Xu2023UniDexGrasp} and UniDexGrasp++ \cite{Wan2023UniDexGraspPP} pushed learning across multiple embodiments, and the MultiGripperGrasp benchmark \cite{casas2024multigrippergrasp} broadened evaluation to more heterogeneous end effectors. More recent work further widened this setting with shared dexterous policies, richer robot-object interaction representations, functional-grasp annotation, and language-aligned grasp supervision \cite{yuan2025crossdex,huang2025resdex,wei2025drograsp,lin2025unifucgrasp,he2025dexvlg}. Once such heterogeneous supervision becomes available, the main question is no longer whether data can be collected, but how one shared generator should represent embodiments whose topology, actuation coupling, and contact geometry differ substantially.

\begin{figure}[t]
    \centering
    \includegraphics[width=\linewidth]{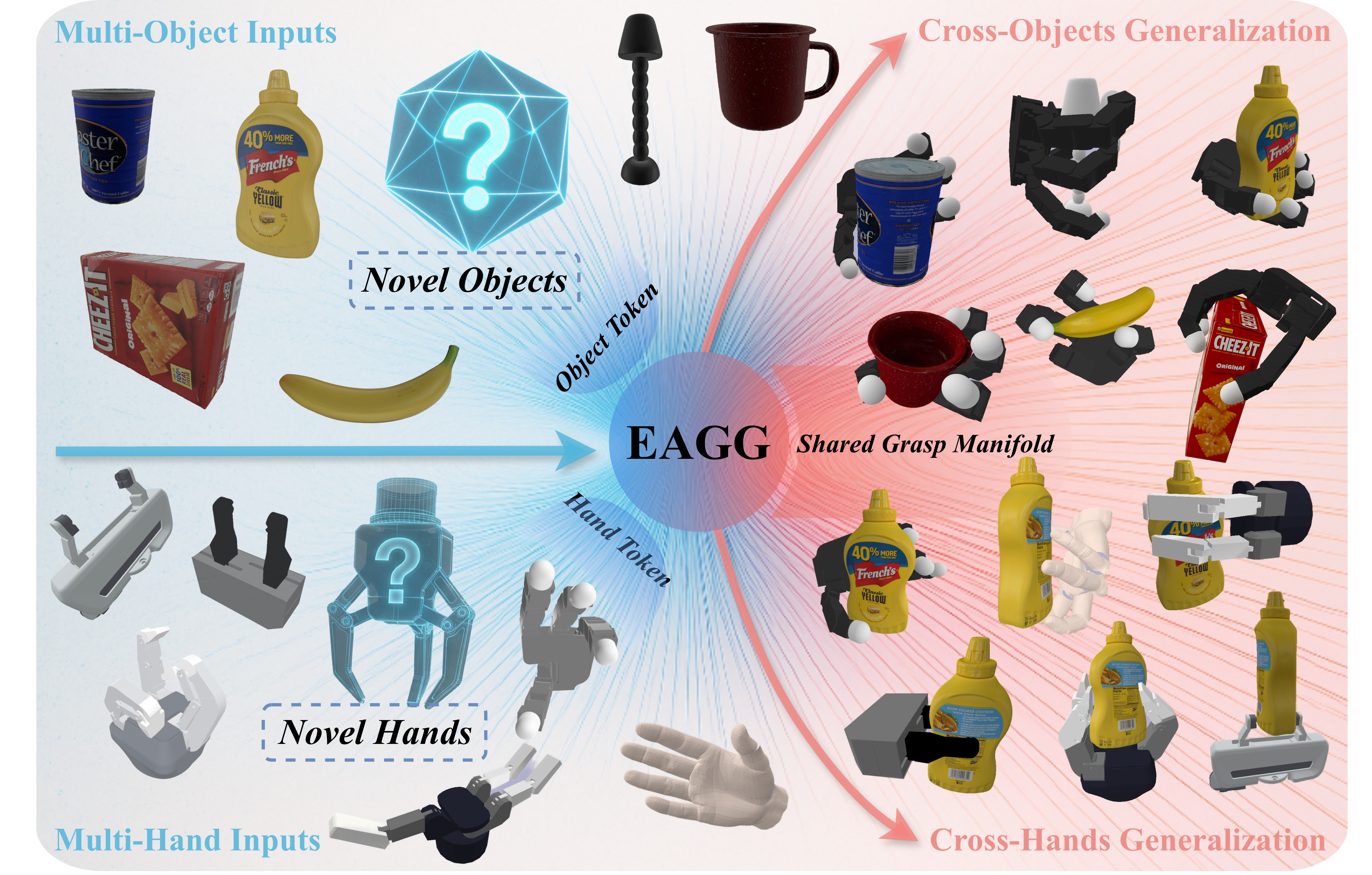}
    \caption{\textbf{Cross-end-effector grasp generation setting.} The figure highlights two coupled generalization axes. Top-left: diverse object inputs, including novel objects. Bottom-left: diverse end effectors, including unseen embodiments with different topology and closure types. Top-right: object generalization for a fixed embodiment, illustrated with Allegro grasps across different objects. Bottom-right: embodiment generalization on the same object. EAGG addresses both axes within one unified generator.}
    \label{fig:teaser}
\end{figure}

Figure~\ref{fig:teaser} summarizes the cross-end-effector grasp generation setting. The left side highlights the diversity of inputs, including novel objects and unseen end effectors with different topology and closure types. The right side then illustrates the two coupled generalization requirements: object generalization for a fixed embodiment and embodiment generalization on the same object. A unified generator must address both axes within one model.

The representation challenge is fundamental. For a fixed embodiment, feasible grasps occupy a structured low-dimensional subset of the full wrist-and-joint space. Across embodiments, those subsets do not collapse into a single common manifold because each end effector has its own joint coordinates, actuation couplings, and closure patterns. Postural-synergy studies \cite{santello1998postural,ciocarlie2007dexterous,Ciocarlie2009IJRR} and robotic embodiment-transfer studies \cite{Gioioso2013SynergyMapping,Santello2016PLR,jiang2021synergy} indicate that grasp postures often admit compact low-dimensional structure without discarding embodiment identity. This motivates an embodiment-aligned representation in which each end effector keeps its own control basis while transfer is carried out in coefficient space.

A compact control space alone is not sufficient. The generator must also know how an end effector is organized and how its geometry changes during grasp formation. A low-dimensional code does not indicate which joints belong to the same kinematic chain, which motions are mechanically coupled, or whether the current state creates emerging collisions or missed contacts. These issues become especially pronounced in cross-end-effector generation because the same nominal code can produce very different articulated geometry on different embodiments.

This difficulty is central to practical deployment. In a shared manipulation stack, swapping between end effectors with different topology or closure behavior should not require relearning grasp generation from scratch, yet forcing all embodiments into one raw joint coordinate system also discards the structure that makes each embodiment effective. A reusable cross-end-effector generator therefore must satisfy two requirements simultaneously: it must preserve embodiment-specific actuation logic, and it must still expose enough common structure for statistical transfer across objects and embodiments. EAGG is designed around this requirement rather than treating embodiment identity as an auxiliary label.

Recent grasp generators synthesize multi-finger grasps directly from object geometry using diffusion \cite{weng2024dexdiffuser}, optimization-guided generative modeling \cite{uria2023grasping}, differentiable simulation \cite{wu2023graspdiff}, or related contact-guided formulations \cite{zhao2024graingrasp,Chen2024SpringGrasp}. Later variants extend this family with uncertainty-aware flow models, task- and language-conditioned generation, and prior-assisted open-set synthesis \cite{feng2025ffhflow,zhang2025dextog,wei2025afforddexgrasp,he2025dexvlg,jian2025gdexgrasp}. In these models, embodiment information is often injected as a static token or latent descriptor. For cross-end-effector generation, static conditioning is restrictive because contact opportunities and collision patterns evolve throughout sampling, especially for embodiments whose articulated geometry changes substantially during closure.

EAGG addresses this problem with embodiment-aligned graph conditioning and Iterative Geometry Injection (IGI). Each embodiment is represented by a topology-aware end-effector graph together with a Principal Component Analysis (PCA)-based low-dimensional control space. A frozen end-effector-cognition backbone converts the current articulated state into geometry-aware point tokens that provide a stable morphology-aware prior, and IGI refreshes those tokens during sampling so that the generator remains synchronized with the evolving end-effector geometry.

The main contributions are:
\begin{enumerate}
    \item \textbf{An embodiment-aligned formulation for cross-end-effector grasp generation.} EAGG represents each embodiment with a topology-aware end-effector graph and a PCA-based low-dimensional control space, enabling one generator to operate across heterogeneous end effectors without imposing a shared raw joint parameterization.
    \item \textbf{Dynamic geometry conditioning through IGI.} EAGG refreshes end-effector conditioning throughout sampling with geometry-aware tokens produced by a frozen end-effector-cognition backbone, providing a stable embodiment prior while allowing the generator to respond to embodiment-specific contact and collision changes as the grasp evolves.
    \item \textbf{An extensive evaluation across jointly trained, finetuned, zero-shot, and real-world regimes.} Across the six jointly trained end effectors, EAGG reaches 56.17\% average success and remains within 1.10 percentage points of specialized training, while IGI reduces the pooled median contact distance from 0.239\,cm to 0.189\,cm and the full framework transfers effectively to held-out end effectors and real-world hardware.
\end{enumerate}

These results show that explicit embodiment alignment supports both cross-object generalization and cross-end-effector transfer within one grasp generator.

\section{Related Work}

\subsection{Fixed-End-Effector Grasp Synthesis and Benchmarks}

Existing methods for grasp synthesis under a fixed embodiment can be broadly grouped into two classes. The first class emphasizes analytic or search-based reasoning, as exemplified by GraspIt! \cite{miller2004graspit} and the Columbia Grasp Database \cite{goldfeder2009columbia}, where grasp quality is tied to explicit geometric, contact, and force-closure analysis. The second class adopts data-driven prediction. Within this class, some methods score sampled candidates, such as Dex-Net \cite{mahler2017dexnet} and PointNetGPD \cite{liang2018pointnetgpd}, while others predict grasp poses more directly from object geometry, such as 6-DoF GraspNet \cite{mousavian20196}, S4G \cite{qin2020s4g}, and Contact-GraspNet \cite{sundermeyer2021contact}. Despite these algorithmic differences, the common setting is unchanged: the end effector is fixed, and the grasp representation is defined around one control space, one contact semantics, and one feasibility region.

This design choice has important consequences for generalization. In fixed-end-effector pipelines, embodiment information is usually absorbed implicitly into the parameterization, training data, and evaluation metric rather than encoded as an explicit transferable variable. Analytic approaches retain strong physical priors and interpretability, but their search spaces become increasingly difficult to manage as articulation grows and sensing becomes noisy. Learning-based methods improve robustness and scalability, yet the learned grasp manifold remains tightly coupled to the embodiment used during training. As a result, these methods often generalize across objects but not across hardware: moving to a new gripper or hand typically requires a new parameterization, new data, or full retraining. The main limitation is not grasp quality per se, but the absence of a unified embodiment representation that would permit cross-end-effector transfer.

Benchmark design evolved in parallel with these methods. YCB \cite{calli2015ycb} and EGAD \cite{morrison2020egad} standardized object diversity and robustness evaluation, while recent reviews \cite{Newbury2023DeepLearningGraspReview,Song2025DexSurvey} documented the transition from isolated grasp detectors to learned synthesis systems. Complementary dataset efforts also explored customizable 6-DoF grasp supervision and visual-tactile stability evaluation \cite{niu2024customizable6dof,niu2026visualtactile}. More recent resources expanded dexterous supervision and embodiment diversity: DexGraspNet \cite{Wang2023DexGraspNet}, DexGraspNet~2.0 \cite{zhang2024dexgraspnet}, UniDexGrasp \cite{Xu2023UniDexGrasp}, and UniDexGrasp++ \cite{Wan2023UniDexGraspPP} increased the scale of dexterous grasp learning, the MGG dataset \cite{casas2024multigrippergrasp} widened evaluation to more heterogeneous end effectors, and later work added functional-grasp and language-aligned supervision for diverse hands \cite{lin2025unifucgrasp,he2025dexvlg}. This progression makes the current gap more visible. Once heterogeneous embodiments appear within one benchmark, the central question becomes whether a single generator can express grasps across them without collapsing their structural differences. EAGG is designed for precisely this regime.

\subsection{Cross-End-Effector Transfer}

Existing approaches to cross-end-effector grasping span four broad strategies. Some use object-centric abstractions, such as contact targets or contact maps, to define a grasp intent that can later be instantiated for different hands \cite{shao2020unigrasp,li2023gengrasp}. Some combine learning with optimization or structured robot-object interaction models so that contact or physics consistency is enforced during adaptation \cite{wang2024transfergrippers,wei2025drograsp,fei2025trograsp}. Others learn policies or action spaces shared across multiple dexterous embodiments \cite{Xu2023UniDexGrasp,Wan2023UniDexGraspPP,yuan2025crossdex,huang2025resdex,yuan2025demograsp}. A final line of work translates grasp distributions more directly between heterogeneous hands \cite{zhong2025grasp2grasp}. These studies show that grasp knowledge need not be learned independently for every hand.

Most current solutions achieve transfer by partially abstracting away the embodiment rather than by defining a unified embodiment representation. Object-centric formulations are attractive because they separate object interaction from embodiment realization, but this separation becomes brittle as morphology diversity grows. A contact plan that is reasonable for a dexterous hand may be difficult to realize for a parallel gripper or an underactuated gripper once closure order, reachable workspace, joint limits, and self-interference are taken into account. Optimization-mediated approaches alleviate some of these issues, but their adaptation cost is shifted to inference time and typically grows with embodiment complexity. Shared-policy dexterous systems reduce hand-specific retraining, yet they are still most naturally suited to settings where the training and test embodiments share broadly similar actuation richness and kinematic structure.

These limitations become sharper in the unseen-end-effector setting. Transfer across embodiments with substantially different topology, closure logic, and contact realizability requires the model to cope with differences not only in geometry, but also in how that geometry can be used during grasp formation. Under such conditions, a coarse morphology token can identify an embodiment without explaining how it is organized or how grasp states should be expressed for it. EAGG addresses this issue by making embodiment structure explicit: grasps are generated through an embodiment-specific low-dimensional control basis and a topology-aware graph, while geometry-aware conditioning is refreshed throughout sampling. In this way, embodiment is not treated as a post-hoc realization constraint, but as a first-class variable in unified grasp generation.

\subsection{Structured Embodiment Representation and Geometry-Aware Generation}

A unified grasp generator for heterogeneous embodiments depends on how the embodiment itself is represented. One line of work uses low-dimensional grasp subspaces or postural synergies to compress high-dimensional articulation into a smaller control interface \cite{santello1998postural,ciocarlie2007dexterous,Ciocarlie2009IJRR}. Related robotic studies \cite{Gioioso2013SynergyMapping,Santello2016PLR,jiang2021synergy} show that such compact representations can preserve actuation structure while improving transfer across dissimilar embodiments. Related work on adaptive synergies and compliant hand design \cite{DellaSantina2018SoftHand2} further shows how embodiment mechanics can encode task-relevant grasp behavior. Their main advantage is that they avoid forcing different hands into one raw joint parameterization. Their limitation is complementary: a low-dimensional code describes how an embodiment may move, but by itself says little about how that embodiment is topologically organized or how its articulated geometry changes during closure. Synergies therefore provide an embodiment-aligned control interface, but not a complete unified embodiment representation.

Another line of work introduces structured morphology through graph-based models \cite{wang2018nervenet,huang2019graph,niu2024customizable6dof}. In robot control more broadly, these representations supply relational inductive bias and enable policies to generalize across varying kinematic trees. For grasp generation, this structural bias is valuable because kinematic chains, joint couplings, and interference patterns are all embodiment dependent. However, graph structure alone is also insufficient. A graph can encode which joints are related, yet it does not define an actuation-aligned coordinate system for expressing grasps across embodiments. Conversely, a compact motion basis can regularize control without preserving the articulated structure needed to reason about contact realization. EAGG combines these two views by treating the control basis and topology-aware graph as complementary components of embodiment representation.

The generation mechanism forms the third ingredient. Diffusion \cite{ho2020denoising}, score-based modeling \cite{song2021score}, flow matching \cite{lipman2023flow,liu2022flow,albergo2023building}, and related continuous-time formulations now provide powerful tools for sampling multimodal continuous actions. In grasping, recent methods have coupled these generative models and related optimization frameworks with object geometry, contact guidance, or optimization signals \cite{wu2023graspdiff,uria2023grasping,weng2024dexdiffuser,zhao2024graingrasp,Chen2024SpringGrasp,Li2023FRoGGeR,Wei2024DexGYS}. Later variants incorporate uncertainty-aware flow modeling, human- or language-guided objectives, and prior-assisted generalization \cite{feng2025ffhflow,huang2025hgdiffuser,zhang2025dextog,wei2025afforddexgrasp,he2025dexvlg,jian2025gdexgrasp}. The shared benefit is stronger geometric awareness and a better ability to model multi-modal grasp distributions than deterministic regression. Yet the trade-offs remain clear. Geometry-heavy optimization can improve physical fidelity, but often incurs higher computational cost and depends on accurate modeling at inference time; purely learned generators are more efficient after training, but can miss embodiment-specific feasibility when conditioning is static or hand-specific. This trade-off is especially problematic for unseen end effectors, where neither fixed hand tokens nor final-pose checks are sufficient. EAGG addresses this gap with a learned flow-matching generator that continuously refreshes geometry-aware conditioning from the evolving articulated state, thereby linking dynamic geometry modeling with a structured embodiment representation.

\section{Method}

\begin{figure*}[t]
    \centering
    \includegraphics[width=0.78\textwidth]{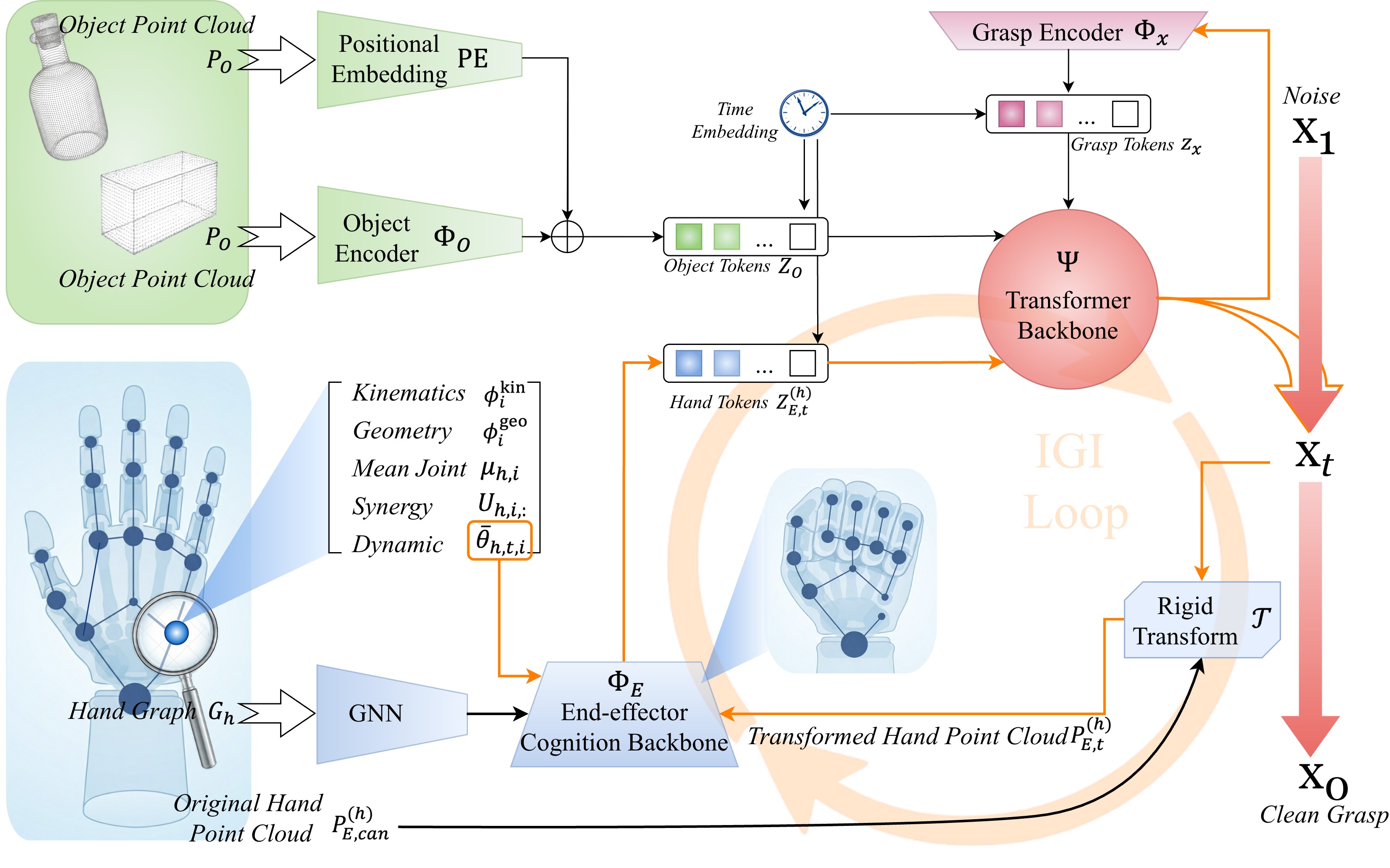}
    \caption{\textbf{EAGG pipeline.} The upper branch encodes the object point cloud into object tokens, while the lower branch encodes the embodiment from a hand graph and a low-dimensional control representation into end-effector tokens. The transformer backbone predicts the grasp state iteratively from noise to the clean grasp. In the IGI loop, the intermediate grasp is decoded to the current articulated geometry, re-encoded by the frozen end-effector-cognition backbone, and fed back as updated end-effector tokens for the next sampling step.}
    \label{fig:architecture}
\end{figure*}

Figure~\ref{fig:architecture} summarizes the full pipeline. EAGG aligns three complementary structures that are tightly coupled in cross-end-effector grasp generation: a low-dimensional control space specifies how an end effector moves, a topology-aware graph specifies how it is organized, and IGI refreshes geometry-aware conditioning from the current sample state. Different embodiments therefore share one generator without being forced into a common raw joint parameterization.

\subsection{Problem Formulation}

Let $P_O \in \mathbb{R}^{N_O \times 3}$ denote the observed object point cloud, and let $h$ denote an end-effector embodiment with actuated joint dimension $q_h$ and kinematic graph $G_h=(V_h,E_h)$. A grasp for embodiment $h$ contains a wrist pose and an embodiment-specific configuration. EAGG predicts the compact grasp state
\begin{equation}
    \mathbf{x} = [\mathbf{s}; \mathbf{t}; \mathbf{r}] \in \mathbb{R}^{d+9},
    \label{eq:grasp_state}
\end{equation}
where $\mathbf{s} \in \mathbb{R}^{d}$ is a low-dimensional control code, $\mathbf{t} \in \mathbb{R}^{3}$ is the wrist translation, and $\mathbf{r} \in \mathbb{R}^{6}$ is a 6-D rotation representation. In all experiments, $d=4$ and the object encoder uses $N_O=1024$ points.

Training data are tuples
\begin{equation}
    \mathcal{D} = \{(P_O^{(n)}, h^{(n)}, \boldsymbol{\theta}_{h^{(n)}}^{(n)}, \mathbf{t}^{(n)}, \mathbf{r}^{(n)})\}_{n=1}^{N},
\label{eq:dataset_tuple}
\end{equation}
where $\boldsymbol{\theta}_{h^{(n)}}^{(n)} \in \mathbb{R}^{q_{h^{(n)}}}$ is the full joint configuration for embodiment $h^{(n)}$. Given $(P_O,h)$, the model outputs $\mathbf{x}$ and decodes it into the wrist pose $(\mathbf{t},\mathbf{r})$ and the embodiment-specific joint configuration $\boldsymbol{\theta}_h(\mathbf{s})$. The central objective is to learn one time-conditioned generator that can operate across embodiments with different kinematic topology, control dimension, and contact geometry.

This parameterization separates two roles that are coupled during grasping but should not be conflated in the representation. The wrist pose captures the global placement of the end effector relative to the object, while the low-dimensional code captures embodiment-specific closure behavior. Such a decomposition is especially useful in the cross-end-effector setting because global object interaction can often be shared at a coarse level even when local articulation differs substantially. The generator can therefore learn transferable object-conditioned approach structure while still decoding the final articulated configuration through an embodiment-aligned interface.

\subsection{Embodiment-Aligned Control Basis and Topology-Aware Graph}

Different end effectors occupy different motion spaces. For each embodiment $h$, feasible grasp postures are collected into a posture matrix $X_h \in \mathbb{R}^{N_h \times q_h}$. An end-effector-specific Principal Component Analysis (PCA) model is then fitted as
\begin{equation}
\begin{aligned}
    \min_{\boldsymbol{\mu}_h, U_h, S_h} \quad &
    \left\| X_h - \mathbf{1}_{N_h}\boldsymbol{\mu}_h^\top - S_h U_h^\top \right\|_F^2 \\
    \text{s.t.} \quad & U_h^\top U_h = I_d,
\end{aligned}
\label{eq:glpca}
\end{equation}
where $\boldsymbol{\mu}_h \in \mathbb{R}^{q_h}$ is the mean posture, $U_h \in \mathbb{R}^{q_h \times d}$ stores the first $d$ principal directions, and $S_h \in \mathbb{R}^{N_h \times d}$ stores the low-dimensional coefficients. This PCA basis defines an embodiment-aligned motion interface: different embodiments correspond to different low-dimensional subspaces, and cross-end-effector transfer is carried out in their coefficient spaces rather than in one shared raw joint vector.

Given a control code $\mathbf{s}$, the full joint configuration is reconstructed by
\begin{equation}
    \boldsymbol{\theta}_h(\mathbf{s}) = \boldsymbol{\mu}_h + U_h \mathbf{s},
    \label{eq:theta_reconstruction}
\end{equation}
and an observed joint configuration can be projected into coefficient space by
\begin{equation}
    \mathbf{s} = U_h^\top (\boldsymbol{\theta}_h - \boldsymbol{\mu}_h).
    \label{eq:theta_projection}
\end{equation}
In implementation, the retained coefficients are standardized with per-end-effector mean and standard deviation before training and unstandardized before applying \eqref{eq:theta_reconstruction}. Each embodiment therefore keeps its own basis $U_h$, while EAGG operates in the coefficient space induced by that basis.

Each embodiment is further represented as a graph derived from its URDF structure. Nodes correspond to the virtual base and actuated joints, and edges connect joints that share local kinematic structure through the same link chain. For node $i$, the static feature vector is defined as
\begin{equation}
    \mathbf{h}^{\mathrm{static}}_{i}
    =
    [\boldsymbol{\phi}_i^{\mathrm{kin}} \parallel \boldsymbol{\phi}_i^{\mathrm{geo}} \parallel \mu_{h,i} \parallel U_{h,i,:}],
    \label{eq:hand_node_static}
\end{equation}
where $\boldsymbol{\phi}_i^{\mathrm{kin}}$ contains joint axis, joint type, limits, local transform, and normalized kinematic depth, $\boldsymbol{\phi}_i^{\mathrm{geo}}$ contains embodiment-level geometric statistics extracted from the URDF, $\mu_{h,i}$ is the mean joint value attached to node $i$, and $U_{h,i,:}$ is the corresponding row of the PCA basis.

At sampling step $t$, the current joint angle reconstructed from the state $\mathbf{x}_t$ is appended to obtain the dynamic node feature
\begin{equation}
    \mathbf{h}_{i,t}^{(0)}
    =
    [\mathbf{h}^{\mathrm{static}}_{i} \parallel \bar{\theta}_{h,t,i}],
    \label{eq:hand_node}
\end{equation}
where $\bar{\theta}_{h,t,i}$ denotes the joint value normalized by its valid range. Graph propagation uses the row-normalized graph convolution
\begin{equation}
    \mathbf{h}_{i,t}^{(\ell+1)}
    =
    \sigma \!\left(
        \mathrm{LN}\!\left(
            \mathbf{h}_{i,t}^{(\ell)}
            + \sum_{j \in \mathcal{N}(i)} \hat{a}_{ij} W_{\ell} \mathbf{h}_{j,t}^{(\ell)}
        \right)
    \right),
    \label{eq:graph_update}
\end{equation}
where $\mathcal{N}(i)$ is the neighbor set of node $i$, $\hat{a}_{ij}$ is the row-normalized adjacency weight derived from $G_h$, $W_{\ell}$ is the learned linear map at layer $\ell$, $\mathrm{LN}(\cdot)$ is layer normalization, and $\sigma(\cdot)$ is the GELU nonlinearity. This stage preserves explicit embodiment structure before the global transformer layers aggregate information across the entire end effector.

The control basis and the graph play complementary roles. The basis specifies which coordinated motions are natural for a given embodiment, while the graph specifies how those motions are distributed over kinematic chains and mechanically coupled parts. Either component alone is incomplete: a basis without topology cannot explain how local actuation is organized, and a graph without a compact control chart still leaves the generator with highly heterogeneous joint spaces. EAGG aligns these two views so that one shared network can reason jointly about embodiment organization and embodiment-specific actuation.

\subsection{Geometry-Aware Dynamic Conditioning}

EAGG uses a frozen end-effector-cognition backbone to convert the current articulated state into geometry-aware point tokens. For embodiment $h$, this module takes as input the topology-aware graph $G_h$, a canonical surface cloud $P_{E,can}^{(h)}$ sampled at the mean posture, and the current articulation induced by the embodiment-specific low-dimensional control state. The graph branch captures kinematic coupling and embodiment-dependent motion directions, while the point branch anchors these cues to concrete surface samples, so each output token carries both local geometric context and the articulation-dependent state of the corresponding region.

This module is pretrained independently from the grasp generator. Its training set is synthesized from URDF meshes, cached surface samples, and the end-effector-specific PCA models. For each embodiment, random low-dimensional codes are drawn, converted back to joint angles through \eqref{eq:theta_reconstruction}, and applied through forward kinematics to obtain posed surface clouds. The network is trained to reconstruct the posed surface from the canonical cloud under graph and articulation conditioning:
\begin{equation}
    \mathcal{L}_{\mathrm{cog}}
    =
    \frac{1}{|P_h|}\sum_{p=1}^{|P_h|}
    \left\|
        \widehat{\mathbf{p}}_{p}
        -
        \mathbf{p}^{\mathrm{pose}}_{p}
    \right\|_2^2,
    \label{eq:cog_loss}
\end{equation}
where $\widehat{\mathbf{p}}_{p}$ and $\mathbf{p}^{\mathrm{pose}}_{p}$ denote the predicted and forward-kinematics-posed coordinates of the $p$-th sampled surface point. After pretraining, the graph encoder and point decoder are loaded into EAGG and frozen during grasp training. Keeping this backbone fixed makes the injected geometry tokens a stable shared morphology prior, prevents grasp supervision from distorting the articulation-to-geometry mapping learned during pretraining, and reduces co-adaptation with the main transformer. Cross-end-effector grasp generalization is then realized downstream when this stable embodiment prior is fused with object geometry and grasp-state evolution inside the shared generator. Figure~\ref{fig:latent_space} later provides a downstream diagnostic of the resulting representation.

Let $P_{E,can}^{(h)}$ denote the canonical cloud of embodiment $h$. At time step $t$, the wrist pose from $\mathbf{x}_t$ transforms it to
\begin{equation}
    \widetilde{P}_{E,t}^{(h)} = \mathcal{T}(\mathbf{t}_t, \mathbf{r}_t)\, P_{E,can}^{(h)},
    \label{eq:igi_geometry}
\end{equation}
where $\mathcal{T}(\cdot)$ applies the current translation and 6-D rotation. The frozen end-effector-cognition backbone then produces dynamic end-effector tokens
\begin{equation}
    Z_{E,t}^{(h)} = \Phi_E\!\left(G_h, \bar{\boldsymbol{\theta}}_{h,t}, \widetilde{P}_{E,t}^{(h)}\right),
    \label{eq:hand_tokens}
\end{equation}
with $\bar{\boldsymbol{\theta}}_{h,t}$ the normalized current joint angles. Each row of $Z_{E,t}^{(h)}$ is a surface-anchored latent that summarizes the local geometry of the current end-effector configuration together with the motion context inherited from $G_h$. During grasp generation, EAGG injects this decoder token stream rather than the explicit posed-cloud output, which preserves a shared geometry-aware representation with modest computational overhead.

Refreshing these tokens online is important because feasibility changes during closure. Early in sampling, the model must infer coarse opposition and enclosure structure, whereas later steps depend more strongly on local contact approach, self-collision, and missed-contact correction. Static embodiment descriptors cannot express this evolution. By recomputing geometry-aware end-effector tokens from the current state, EAGG exposes the generator to the articulated geometry that will actually determine whether the current trajectory can converge to an executable grasp.

The object stream is encoded as
\begin{equation}
    Z_O = \Phi_O(P_O) + \operatorname{PE}(P_O),
    \label{eq:object_tokens}
\end{equation}
where $\Phi_O$ is a PointNet++-style point-cloud backbone \cite{Qi2017PointNetplusplus} and $\operatorname{PE}(\cdot)$ is a positional embedding derived from normalized point coordinates.

EAGG fuses object and end-effector conditioning through three token families: a single grasp-state token, dynamic end-effector tokens, and object point tokens. Let $z_x=\Phi_x(\mathbf{x}_t)\in\mathbb{R}^{D}$ denote the encoded grasp token, let $Z_{E,t}^{(h)} \in \mathbb{R}^{P_h \times D}$ and $Z_O \in \mathbb{R}^{M_O \times D}$ denote the end-effector and object token matrices, let $e_t = E_t(t)\in\mathbb{R}^{D}$ be the sinusoidal time embedding, and let $e_g,e_e,e_o \in \mathbb{R}^{D}$ be learned type embeddings for the grasp, end-effector, and object streams. The token sequence passed to the main transformer is
\begin{equation}
    \widetilde{Z}_t
    =
    \left[
    \begin{array}{c}
        z_x + e_t + e_g \\
        Z_{E,t}^{(h)} + \mathbf{1}_{P_h}(e_t + e_e)^\top \\
        Z_O + \mathbf{1}_{M_O}(e_t + e_o)^\top
    \end{array}
    \right],
    \label{eq:token_fusion}
\end{equation}
where $\mathbf{1}_{P_h}$ and $\mathbf{1}_{M_O}$ broadcast the global time-and-type condition over the end-effector and object tokens. The first output token predicts the clean grasp state,
\begin{equation}
    \widehat{\mathbf{x}}_0
    =
    W_{\mathrm{out}}
    \big[\Psi(\widetilde{Z}_t)\big]_{\mathrm{grasp}},
    \label{eq:transformer_head}
\end{equation}
with $\Psi(\cdot)$ the transformer backbone and $[\cdot]_{\mathrm{grasp}}$ selecting the updated grasp token. This design fuses low-dimensional control, embodiment structure, dynamic geometry, and object geometry inside one interaction backbone.

\subsection{Training Objective and Iterative Geometry Injection}

During training, a clean grasp state $\mathbf{x}_0$ is interpolated with Gaussian noise $\boldsymbol{\epsilon} \sim \mathcal{N}(\mathbf{0}, I_{d+9})$ along the linear path
\begin{equation}
    \mathbf{x}_t = (1-t)\mathbf{x}_0 + t\boldsymbol{\epsilon}, \qquad t \in [0,1].
    \label{eq:flow_path}
\end{equation}
IGI makes the conditioning explicitly state dependent by recomputing the end-effector tokens from the current state at each sampled time. The time-varying conditioning bundle is
\begin{equation}
    \mathcal{C}_t = \{Z_O, Z_{E,t}^{(h)}\},
    \label{eq:dynamic_condition}
\end{equation}
where $\mathcal{C}_t$ contains the object tokens and the geometry-aware end-effector tokens associated with the current noisy grasp state. The network predicts the clean state directly,
\begin{equation}
    \widehat{\mathbf{x}}_0 = f_\Theta(\mathbf{x}_t, t, \mathcal{C}_t).
    \label{eq:vector_field}
\end{equation}

We optimize a component-wise Huber objective on the reconstructed clean state,
\begin{equation}
\begin{aligned}
    \mathcal{L}
    =
    \mathbb{E}
    \Big[\omega(t) (&\lambda_{\mathrm{code}}\, \ell_{\mathrm{H}}(\widehat{\mathbf{s}}_0, \mathbf{s}_0)
    + \lambda_{\mathrm{pos}}\, \ell_{\mathrm{H}}(\widehat{\mathbf{t}}_0, \mathbf{t}_0) \\
    &+ \lambda_{\mathrm{rot}}\, \ell_{\mathrm{H}}(\widehat{\mathbf{r}}_0, \mathbf{r}_0))\Big],
\end{aligned}
    \label{eq:overall_loss}
\end{equation}
where the expectation is taken over $(P_O,h,\boldsymbol{\theta}_h,\mathbf{t}_0,\mathbf{r}_0)\sim\mathcal{D}$, time $t$, and noise $\boldsymbol{\epsilon}$, $\mathbf{s}_0$ is obtained from $\boldsymbol{\theta}_h$ via \eqref{eq:theta_projection}, and $\mathbf{x}_0=[\mathbf{s}_0;\mathbf{t}_0;\mathbf{r}_0]$. The loss $\ell_{\mathrm{H}}(\cdot,\cdot)$ denotes the average Huber loss over the corresponding dimensions, and $\omega(t)=\exp(-2t)$ emphasizes lower-noise states. The low-dimensional code term can additionally be reweighted component-wise to emphasize leading PCA directions.

This objective is matched to the intended sampling behavior. Lower-noise states receive greater emphasis because the final phase of generation is where embodiment-specific contact geometry matters most, and errors in this regime are more likely to translate into penetration, missed closure, or unstable grasp ordering. In effect, the loss encourages the model not only to recover the coarse grasp family, but also to resolve the final articulated state with enough precision for executable contact formation across heterogeneous embodiments.

At inference time, IGI recomputes $Z_{E,t}^{(h)}$ after every update. Given the predicted clean state at time $t$, the corresponding noise estimate is recovered as
\begin{equation}
    \widehat{\boldsymbol{\epsilon}}_t
    =
    \frac{\mathbf{x}_t - (1-t)\widehat{\mathbf{x}}_0}{t},
    \label{eq:inference_geometry}
\end{equation}
and the next state on the decreasing time schedule is reconstructed by
\begin{equation}
    \mathbf{x}_{t'}
    =
    (1-t')\widehat{\mathbf{x}}_0 + t'\widehat{\boldsymbol{\epsilon}}_t,
    \qquad 0 \le t' < t.
    \label{eq:inference_update}
\end{equation}
When IGI is disabled, the dynamic angle term in \eqref{eq:hand_node} is frozen at the mean posture, which yields the Non-IGI reference used later in Fig.~\ref{fig:igi_diagnostic}.

\section{Experiments}
\label{sec:experiments}

Experiments evaluate whether one generator can preserve grasp quality across heterogeneous end effectors while benefiting from embodiment alignment and IGI. We first compare EAGG with representative baselines, then analyze performance across training, finetuning, and zero-shot end effectors, followed by transfer diagnostics, ablations, efficiency evaluation, and real-world execution.

\subsection{Experimental Setup}

Experiments use the \emph{MultiGripperGrasp} (MGG) dataset \cite{casas2024multigrippergrasp}. The object split reserves 50 objects for testing, and these test objects are excluded from the base-training object set. After filtering grasp candidates with fall time $\tau_{h,o,k}\geq 3$\,s, the effective base-training and test pools are
\begin{equation}
\begin{aligned}
    \mathcal{D}_{\mathrm{base}} &= \{(h,o,k) \mid h \in \mathcal{H}_{\mathrm{train}},\; o \in \mathcal{O}_{\mathrm{train}},\; \tau_{h,o,k} \geq 3\,\mathrm{s}\}, \\
    \mathcal{D}_{\mathrm{test}} &= \{(h,o,k) \mid h \in \mathcal{H}_{\mathrm{all}},\; o \in \mathcal{O}_{\mathrm{test}},\; \tau_{h,o,k} \geq 3\,\mathrm{s}\},
\end{aligned}
\label{eq:dataset_split}
\end{equation}
where $\mathcal{H}_{\mathrm{train}}$ is the jointly trained end-effector set, $\mathcal{H}_{\mathrm{all}}$ contains all evaluated end effectors, and $\mathcal{O}_{\mathrm{train}}$ and $\mathcal{O}_{\mathrm{test}}$ denote the training and held-out object sets.

The evaluated embodiments are divided into three categories. The \emph{training end-effector set} contains Allegro, Barrett, Franka Panda, Robotiq~3F, WSG-50, and HumanHand. EAGG base training is carried out jointly on all six training end effectors within one shared model. The \emph{finetuning end-effector set} contains Sawyer and Jaco; both appear in MGG but are excluded from joint base training and adapted afterward for 10 epochs using 5\% of their grasp data. The \emph{zero-shot end-effector set} contains FreedomHand and DexHand. For these two embodiments, a small seed set of grasps is synthesized on basic objects with SynergyGrasp \cite{niu_synergygrasp_2026} and then used for lightweight adaptation.

Unless otherwise stated, EAGG uses a 256-dimensional embedding width, 8 attention heads, 8 network blocks, a 4-dimensional low-dimensional control code, a batch size of 420, and 10 base-training epochs. Optimization uses Adam with learning rate $2\times10^{-4}$ and zero weight decay, and the sampled time horizon increases from $t_{\max}=0.3$ to $0.98$ over the first five epochs. Wrist translations are scaled by 10 during optimization and mapped back to metric units during evaluation. All training and evaluation experiments are run on a workstation with 4 NVIDIA RTX 4090 GPUs.

Table~\ref{tab:main_results} uses the following baseline abbreviations. \emph{NS} denotes normal-aligned sampling, a heuristic baseline reported for WSG-50. \emph{GPG} denotes Grasp Pose Generator \cite{tenPas2017GPD}. \emph{UDG} denotes UniDexGrasp \cite{Xu2023UniDexGrasp}. \emph{DD} denotes the DexDiffusion family based on DexDiffuser \cite{weng2024dexdiffuser}, with variants \emph{DD (pn2)}, \emph{DD (bps)}, \emph{DD (bps+EGD)}, and \emph{DD (2stage)}. Component-level EAGG variants are reported separately in Table~\ref{tab:ablation}.

We report success rate (SR, \%), contact distance (CD, cm), penetration depth (PD, cm), contact count (CC), and repeated-grasp ratio (RGR, \%). Higher SR and CC are better, while lower CD, PD, and RGR are better. SR measures task completion, CD and PD measure geometric consistency, CC measures contact richness, and RGR captures grasp diversity.

This protocol separates three sources of difficulty that are often conflated: object novelty, embodiment novelty, and adaptation budget. The training set measures whether one generator can share grasp knowledge without discarding embodiment-specific control structure. The finetuning set measures how quickly that shared prior can be specialized when a small amount of embodiment data becomes available. The zero-shot set measures whether the same architecture can bootstrap to end effectors outside the original joint training pool through lightweight seed grasps.

For the training-end-effector analysis, we report both the unified model and a specialized counterpart with the same architecture trained on a single embodiment. This comparison measures the cost of unification directly, rather than comparing a generalist model only against heterogeneous external baselines. A small unified-to-specialized gap indicates that cross-end-effector sharing is capturing reusable structure instead of simply averaging away embodiment-specific behavior.

\subsection{Cross-End-Effector Benchmark Results}

The benchmark stresses two transfer axes simultaneously. Along the object axis, test objects are excluded from the base-training pool. Along the embodiment axis, the model must span jointly trained, finetuned, and lightweight-adapted end effectors without reverting to one model per embodiment. This setting reveals whether a shared generator can reuse statistical strength across morphologies while preserving embodiment-specific structure.

Table~\ref{tab:main_results} compares EAGG with representative baselines on three training end effectors that cover markedly different closure mechanisms: WSG-50, Robotiq~3F, and Allegro. The table reports the jointly trained model \emph{EAGG (Unified)} and an end-effector-specific model \emph{EAGG (Specialized)}. Ablation variants are deferred to Table~\ref{tab:ablation} so that the main comparison remains focused on external baselines.

The quantitative pattern is consistent across all three embodiments. Relative to the strongest non-EAGG baseline, EAGG (Unified) improves SR by 4.22 points on WSG-50, 17.70 points on Robotiq~3F, and 35.46 points on Allegro. The gain grows as the closure mechanism becomes more articulated: WSG-50 is constrained by nearly one-dimensional parallel-jaw motion, Robotiq~3F benefits from coordinated adaptive fingers, and Allegro exposes the largest morphology mismatch and the largest gain from explicit embodiment alignment.

The gap between the unified and specialized models remains small. EAGG (Specialized) exceeds EAGG (Unified) by 1.72 points on WSG-50, 0.84 points on Robotiq~3F, and 1.13 points on Allegro. Shared training therefore preserves most of the attainable task success while embodiment-specific specialization mainly sharpens the final contact state.

Several baselines remain competitive on individual proxy metrics. DD (pn2) attains the best CD and PD on WSG-50 and Robotiq~3F, while DD (bps+EGD) yields the highest CC on Robotiq~3F and Allegro among the external baselines. Table~\ref{tab:main_results} therefore should be read as evidence of stronger end-to-end grasp completion rather than uniform dominance on every proxy metric.

The diversity metric provides a complementary interpretation. RGR remains near zero for Allegro and HumanHand, but rises sharply for the hardest parallel-jaw cases, especially WSG-50. When the feasible closure family is narrow, a unified generator tends to revisit similar solutions even when success rate improves. Embodiments with richer contact options preserve both stronger success and stronger diversity.

Table~\ref{tab:main_results} also shows that success rate and local geometry proxies need not move in lockstep across embodiments. On Allegro, EAGG gains a large SR margin while remaining only moderately different from the strongest baselines on CD and PD. For highly articulated end effectors, the main challenge is not merely to minimize local penetration or distance, but to land in a kinematically feasible contact arrangement with the right finger ordering and enclosure pattern. On WSG-50, by contrast, small changes in CD or PD translate more directly into success or failure because the closure space is so narrow.

Table~\ref{tab:main_results} suggests that the main advantage of EAGG is structural rather than merely numerical. The model does not dominate every proxy metric on every embodiment, yet it consistently produces the strongest task-level completion once morphology diversity becomes meaningful. This is precisely the regime targeted by embodiment alignment: the goal is to preserve the structural information needed to realize feasible grasps on heterogeneous end effectors, not simply to optimize one local geometric statistic in isolation.

\begin{table*}[t]
\caption{Representative comparison on three training end effectors.}
\label{tab:main_results}
\centering
\footnotesize
\setlength{\tabcolsep}{2.6pt}
\renewcommand{\arraystretch}{0.95}
\begin{tabular}{llccccc}
\hline
\textbf{End-effector} & \textbf{Method} & \textbf{SR (\%)} & \textbf{CD (cm)} & \textbf{PD (cm)} & \textbf{CC} & \textbf{RGR (\%)} \\
\hline
\multirow{9}{*}{WSG-50}
& NS & 5.33 & 0.45 & 1.10 & 0.790 & 0.10 \\
& GPG & 10.28 & 0.45 & 0.80 & 1.203 & 4.50 \\
& UDG & 4.55 & 0.48 & 0.52 & 0.618 & 0.00 \\
& DD (pn2) & 4.32 & 0.28 & 0.35 & 0.725 & 0.60 \\
& DD (bps) & 9.34 & 0.36 & 0.45 & 1.186 & 3.00 \\
& DD (bps+EGD) & 9.28 & 0.37 & 0.45 & 1.185 & 3.10 \\
& DD (2stage) & 9.26 & 0.37 & 0.45 & 1.184 & 3.10 \\
& EAGG (Unified) & 14.50 & 0.40 & 0.77 & 1.262 & 8.76 \\
& EAGG (Specialized) & \best{16.22} & 0.33 & 0.64 & 1.817 & 1.73 \\
\hline
\multirow{7}{*}{Robotiq 3F}
& UDG & 30.00 & 0.44 & 0.58 & 1.769 & 0.00 \\
& DD (bps) & 35.72 & 0.45 & 0.62 & 1.966 & 0.80 \\
& DD (pn2) & 57.54 & 0.35 & 0.53 & 2.701 & 0.30 \\
& DD (bps+EGD) & 60.72 & 0.37 & 0.56 & 2.888 & 0.30 \\
& DD (2stage) & 60.55 & 0.37 & 0.56 & 2.881 & 0.30 \\
& EAGG (Unified) & 78.42 & 1.14 & 1.48 & 2.512 & 1.20 \\
& EAGG (Specialized) & \best{79.26} & 0.78 & 1.07 & 2.615 & 0.18 \\
\hline
\multirow{7}{*}{Allegro}
& UDG & 22.73 & 0.51 & 0.60 & 1.922 & 0.00 \\
& DD (bps) & 26.74 & 0.45 & 0.58 & 2.496 & 0.04 \\
& DD (pn2) & 50.74 & 0.43 & 0.59 & 4.092 & 0.10 \\
& DD (bps+EGD) & 50.94 & 0.44 & 0.60 & 4.129 & 0.14 \\
& DD (2stage) & 51.00 & 0.44 & 0.60 & 4.125 & 0.14 \\
& EAGG (Unified) & 86.46 & 0.50 & 1.04 & 2.564 & 0.12 \\
& EAGG (Specialized) & \best{87.59} & 0.41 & 0.91 & 2.685 & 0.00 \\
\hline
\end{tabular}
\end{table*}
\renewcommand{\arraystretch}{1.0}

Table~\ref{tab:cross_morphology} extends the picture to all 10 end effectors and reveals a clear regime structure. Within the training set, performance spans from 14.50\% SR on WSG-50 to 86.46\% on Allegro, so transfer difficulty is non-uniform even for jointly trained embodiments. Across the six training end effectors as a whole, however, EAGG (Unified) reaches 56.17\% SR, only 1.10 points below the 57.27\% average of specialized training.

That geometry gap is visible in the averaged metrics. Specialized training reduces CD from 0.56 to 0.40\,cm and PD from 1.04 to 0.79\,cm, while increasing CC from 2.081 to 2.254. Embodiment-specific specialization therefore mainly improves how the end effector settles onto the object and how contact is formed in the final grasp.

The held-out regimes provide the stronger transfer test. Finetuning end effectors reach 36.68\% average SR after only 10 epochs with 5\% data, which is 19.49 points below the training average while already demonstrating fast embodiment-specific adaptation. The zero-shot end effectors reach 33.25\% average SR after lightweight adaptation from SynergyGrasp seed grasps \cite{niu_synergygrasp_2026}, only 3.43 points below the finetuning average despite not belonging to the original joint training pool. Jaco adapts more successfully than Sawyer, and DexHand slightly exceeds FreedomHand in SR while FreedomHand preserves richer contact statistics.

Performance across embodiments also exposes the main physical boundary of the benchmark. Less articulated systems such as Franka Panda and WSG-50 remain the hardest because many test objects are too large for shape closure, so success depends more strongly on force closure under strict simulator settings. End effectors with richer articulation, by contrast, can redistribute contact across multiple joints and recover from small approach errors more effectively.

The per-end-effector rows also show that performance is not determined by DoF count alone. Barrett attains 80.62\% SR despite having far fewer articulated degrees of freedom than HumanHand, whereas HumanHand retains richer contact statistics. Likewise, the low RGR of Allegro and HumanHand compared with Franka Panda and WSG-50 indicates that grasp diversity is easier to preserve when the embodiment can realize several distinct closure patterns on the same object.

These regime differences are informative for model behavior. The training regime measures whether one shared generator can retain embodiment structure under joint optimization, the finetuning regime measures how efficiently that structure can be specialized, and the zero-shot regime measures whether lightweight seeding provides a useful starting point for previously unseen embodiments. EAGG performs coherently across all three regimes, which indicates that the learned representation is reusable not only at convergence but also during adaptation.

\begin{table*}[t]
\caption{EAGG performance across training, finetuning, and zero-shot end-effectors.}
\label{tab:cross_morphology}
\centering
\footnotesize
\setlength{\tabcolsep}{3.5pt}
\begin{tabular}{llccccc}
\hline
\textbf{Split} & \textbf{End-effector} & \textbf{SR (\%)} & \textbf{CD (cm)} & \textbf{PD (cm)} & \textbf{CC} & \textbf{RGR (\%)} \\
\hline
\multirow{7}{*}{Training (Unified)}
& Allegro & 86.46 & 0.50 & 1.04 & 2.564 & 0.12 \\
& Barrett & 80.62 & 0.36 & 0.98 & 1.709 & 0.34 \\
& Franka Panda & 22.87 & 0.37 & 0.81 & 1.991 & 8.54 \\
& Robotiq 3F & 78.42 & 1.14 & 1.48 & 2.512 & 1.20 \\
& WSG-50 & 14.50 & 0.40 & 0.77 & 1.262 & 8.76 \\
& HumanHand & 54.18 & 0.60 & 1.15 & 2.450 & 0.00 \\
& Average & 56.17 & 0.56 & 1.04 & 2.081 & 3.16 \\
\hline
\multirow{7}{*}{Training (Specialized)}
& Allegro & 87.59 & 0.41 & 0.91 & 2.685 & 0.00 \\
& Barrett & 81.53 & 0.26 & 0.75 & 1.770 & 0.03 \\
& Franka Panda & 23.88 & 0.21 & 0.62 & 2.107 & 5.33 \\
& Robotiq 3F & 79.26 & 0.78 & 1.07 & 2.615 & 0.18 \\
& WSG-50 & 16.22 & 0.33 & 0.64 & 1.817 & 1.73 \\
& HumanHand & 55.16 & 0.40 & 0.78 & 2.531 & 0.00 \\
& Average & 57.27 & 0.40 & 0.79 & 2.254 & 1.21 \\
\hline
\multirow{3}{*}{Finetuning}
& Sawyer & 29.21 & 0.59 & 1.05 & 1.737 & 7.01 \\
& Jaco & 44.14 & 0.13 & 0.19 & 2.655 & 0.25 \\
& Average & 36.68 & 0.36 & 0.62 & 2.196 & 3.63 \\
\hline
\multirow{3}{*}{Zero-shot}
& FreedomHand & 31.83 & 0.55 & 0.71 & 2.774 & 0.00 \\
& DexHand & 34.67 & 0.56 & 0.62 & 1.890 & 0.00 \\
& Average & 33.25 & 0.56 & 0.67 & 2.332 & 0.00 \\
\hline
\end{tabular}
\end{table*}

\subsection{Transfer Dynamics and Representation Structure}

\begin{figure}[!t]
    \centering
    \includegraphics[width=\linewidth,height=0.58\textheight,keepaspectratio]{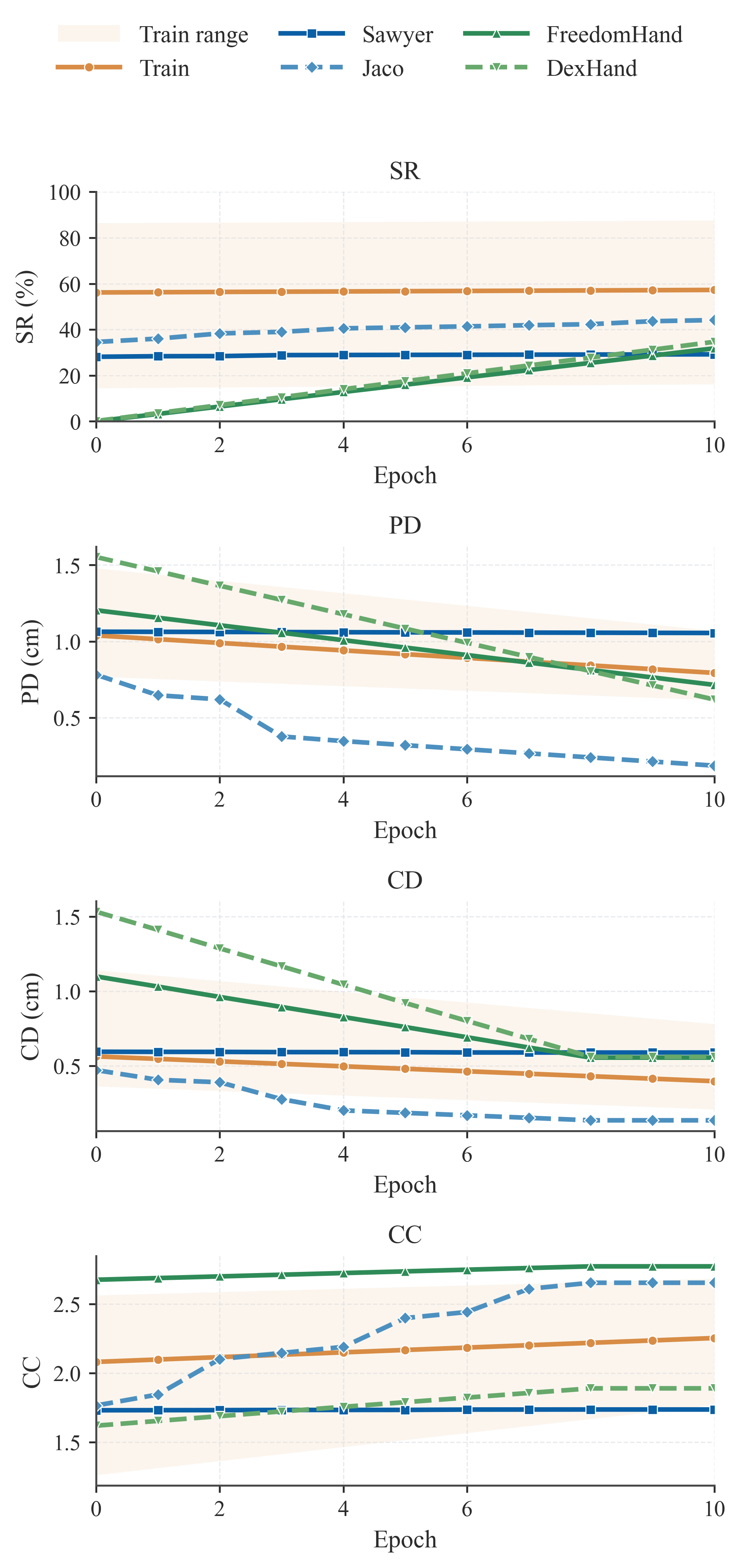}
    \caption{\textbf{Cross-end-effector adaptation trends.} Curves compare the training-end-effector average, two finetuning end effectors (Sawyer and Jaco), and two zero-shot end effectors (FreedomHand and DexHand) over 10 adaptation epochs. The shaded band marks the min--max envelope of the six training end effectors.}
    \label{fig:finetune_efficiency}
\end{figure}

Figure~\ref{fig:finetune_efficiency} resolves Table~\ref{tab:cross_morphology} into epoch-wise adaptation trajectories. Jaco improves fastest: most of its SR gain appears in the early epochs, and its CD and PD curves fall quickly toward low-error values. Sawyer also benefits from lightweight adaptation but improves more gradually, which indicates that adaptation efficiency depends on morphology compatibility rather than on the mere availability of a small calibration set. FreedomHand and DexHand both improve steadily from low initial performance, but their trajectories remain more sensitive to embodiment-specific contact ordering, especially in CC.

The shaded training-end-effector envelope is useful for interpretation. Jaco approaches the lower edge of the training regime within ten epochs, whereas Sawyer and the two zero-shot end effectors remain outside that envelope on some metrics even when their SR rises substantially. The main empirical pattern is that SR can improve relatively quickly, while fine contact geometry and contact richness usually take longer to move toward the training-end-effector distribution.

\begin{figure}[!t]
    \centering
    \includegraphics[width=\linewidth,height=0.5\textheight,keepaspectratio]{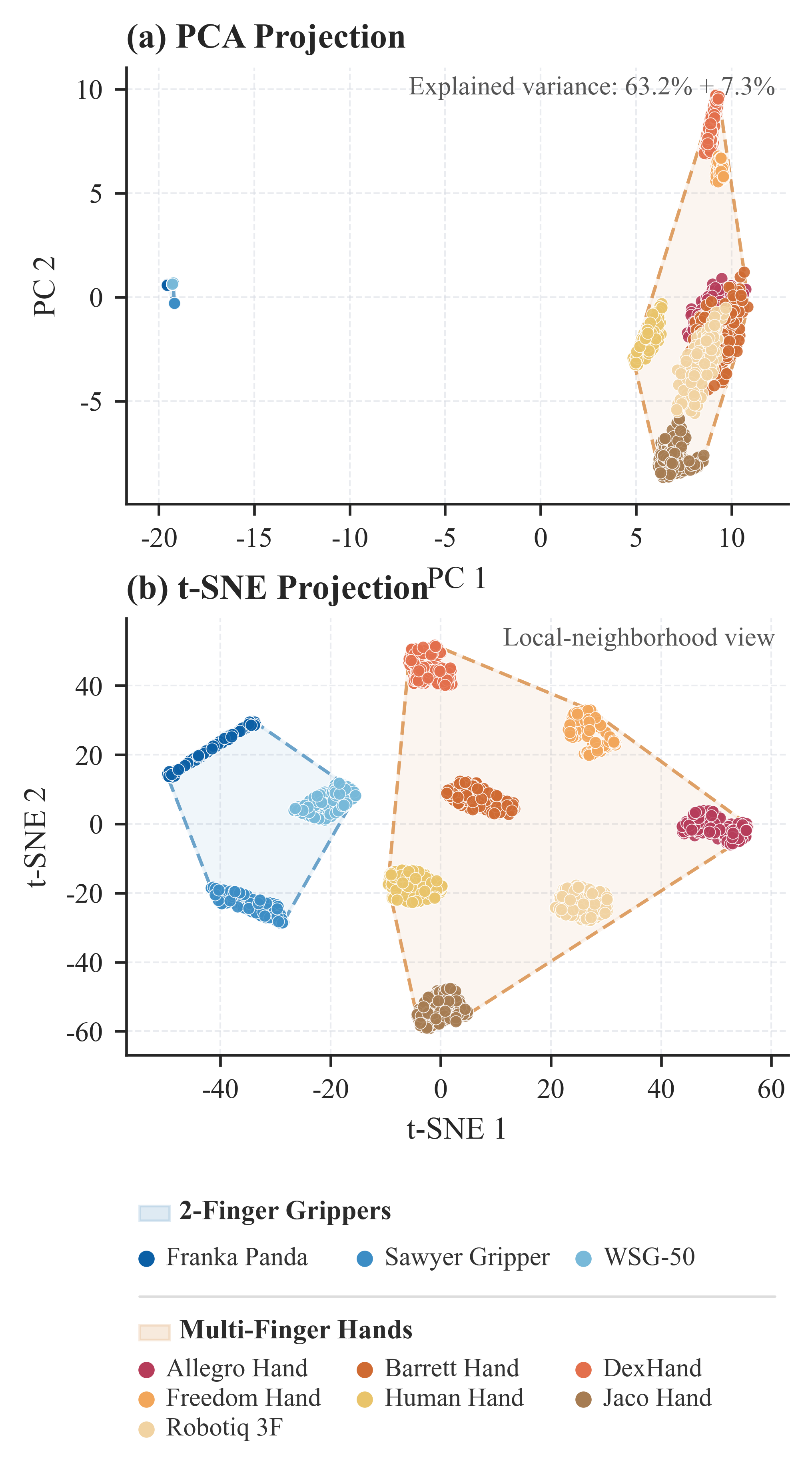}
    \caption{\textbf{Latent-space diagnostic.} Projection of end-effector embeddings exported from the pre-trained end-effector-cognition model. The figure provides supporting evidence that the representation preserves morphology-relevant organization.}
    \label{fig:latent_space}
\end{figure}

Figure~\ref{fig:latent_space} provides supporting evidence for the representation component of this transfer behavior. The exported end-effector-cognition embeddings separate two-finger grippers from multi-finger embodiments and preserve structure within each group rather than collapsing all embodiments into one undifferentiated cluster. Quantitatively, the mean between-end-effector centroid distance is 16.41, whereas the mean within-end-effector spread is only 0.57, yielding a separation ratio of about 29:1.

This morphology-aware organization is important for transfer. Adaptation does not start from a representation that must rediscover embodiment identity from scratch; instead, the shared model already arranges end effectors in a space that reflects topology and closure behavior. This reduces the burden on downstream adaptation, while task-level transfer still depends on how these tokens interact with object features and grasp-state updates inside the shared generator. It also helps explain why small amounts of data can already improve task success, even though contact geometry and local control statistics may continue to adjust afterward.

The projection is also informative at a coarser semantic level. Two-finger grippers occupy a compact region separated from the broader manifold of multi-finger embodiments, while embodiments within each family remain distinguishable. This is the behavior desired from a transferable end-effector representation: it should preserve large morphology classes strongly enough to guide adaptation, but it should not collapse different embodiments inside a class into identical tokens. The latent plot therefore supports the interpretation that EAGG shares structure across related embodiments without erasing the distinctions needed for embodiment-specific closure.

Taken together, the adaptation traces and the latent projection suggest that EAGG provides a structured starting point for new embodiments rather than leaving each one to be learned from scratch. The most reliable observation is that success rate often improves earlier than CC, CD, and PD, indicating that executable grasps can emerge before contact geometry is fully calibrated. This interpretation is consistent with the small gap between unified and specialized training on the training end effectors.

\subsection{Geometry Diagnostics and Component Analysis}

\begin{figure}[!t]
    \centering
    \includegraphics[width=\linewidth,height=0.58\textheight,keepaspectratio]{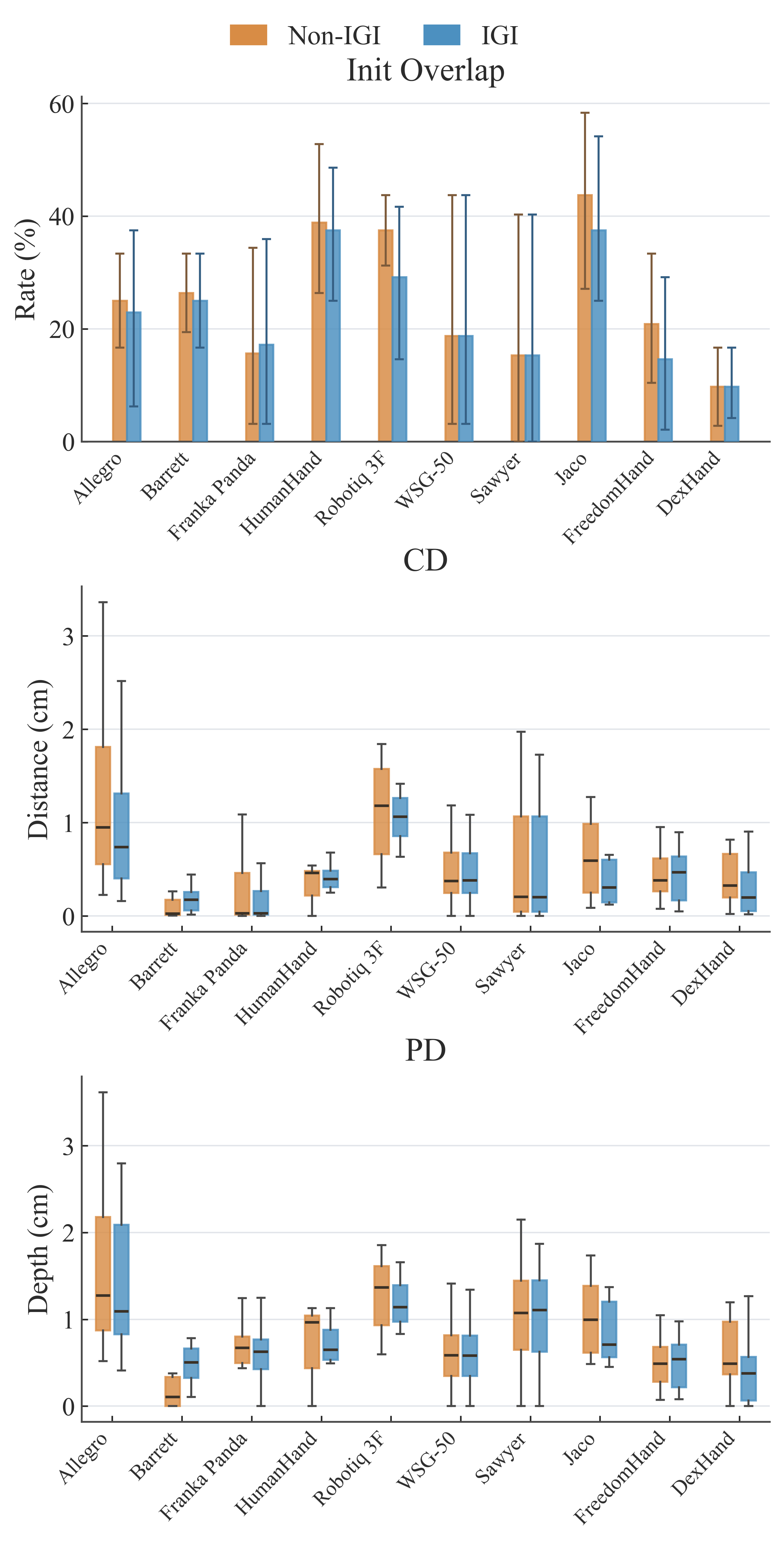}
    \caption{\textbf{Cross-end-effector IGI diagnostic.} Orange and blue boxplots compare Non-IGI and IGI on 10 end effectors. The three panels report initial overlap (\%), final contact distance (CD, cm), and final penetration depth (PD, cm); lower values are better in all cases. Boxplots aggregate per-object summaries, so the figure measures whether IGI consistently shifts the geometry distribution toward cleaner states across embodiments.}
    \label{fig:igi_diagnostic}
\end{figure}

Figure~\ref{fig:igi_diagnostic} isolates the effect of IGI on geometry quality across all 10 end effectors. When the per-object summaries are aggregated over embodiments, IGI lowers the median CD from 0.406\,cm to 0.350\,cm and the median PD from 0.731\,cm to 0.700\,cm; the mean initial overlap also decreases from 24.67\% to 22.46\%. The final state is therefore cleaner, and the improvement appears both at initialization and at the end of sampling. IGI therefore steers the trajectory toward more consistent end-effector-object configurations rather than merely postponing collision problems to later steps.

The embodiment breakdown is equally informative. Allegro, Jaco, Robotiq~3F, and HumanHand show the clearest downward shifts in CD and PD, whereas Franka Panda and WSG-50 are close to neutral. This differential effect matches the mechanism: when an embodiment can substantially change its articulated geometry during closure, refreshing the end-effector condition online changes the effective contact landscape at each step. For almost one-dimensional parallel-jaw motion, there is simply less state-dependent geometry for IGI to exploit.

Table~\ref{tab:ablation} shows that IGI is one part of a broader component picture. The two largest SR drops come from removing topology-aware graph conditioning (56.17\% to 37.77\%) and removing end-effector geometry (56.17\% to 38.75\%), which identifies morphology structure and dynamic geometry as the primary ingredients of the model. Removing the basis prior also matters, reducing SR to 43.48\% and worsening both CD and PD. The low-dimensional control interface therefore does more than compress the posture space; it helps align heterogeneous embodiments into a reusable control representation.

The remaining ablations show the supporting roles of the other cues. Removing local object features or absolute pose lowers SR by more than 14 points, so object conditioning matters throughout sampling rather than only at initialization. The \emph{w/o Flow} variant even reduces RGR relative to full EAGG, yet its SR, CD, and CC all worsen, indicating that no single proxy metric alone captures executable grasp quality. The ablations suggest a clear division of labor: the basis prior provides a reusable control interface, graph conditioning injects embodiment structure, and IGI keeps the generator synchronized with the evolving articulated geometry.

The per-end-effector drops reinforce that this is not a narrow effect limited to one embodiment class. Without the graph prior, Barrett falls from 80.62\% SR to 43.21\%, HumanHand drops from 54.18\% to 22.59\%, and Franka Panda drops from 22.87\% to 9.25\%. These examples span end effectors with different articulation patterns and control complexity, indicating that topology-aware conditioning is a general mechanism for making the generator respect embodiment-specific actuation structure.

The ablation trends also help explain why EAGG generalizes better than a purely static embodiment token. Removing graph conditioning or end-effector geometry does not merely reduce average performance; it alters the failure mode of the generator. Predictions become less consistent in closure ordering, less stable in final contact formation, and more sensitive to morphology-specific ambiguities. The full model is therefore benefiting from a structured interaction between the compact control basis, topology-aware message passing, and state-dependent geometry updates, rather than from any single cue in isolation.

\begin{table}[t]
\caption{Average ablation results on the six training end effectors.}
\label{tab:ablation}
\centering
\footnotesize
\setlength{\tabcolsep}{3.3pt}
\begin{tabular}{lccccc}
\hline
\textbf{Variant} & \textbf{SR (\%)} & \textbf{CD (cm)} & \textbf{PD (cm)} & \textbf{CC} & \textbf{RGR (\%)} \\
\hline
Full EAGG & \best{56.17} & \best{0.56} & \best{1.04} & \best{2.081} & 3.16 \\
w/o LocalFeat & 41.67 & 1.03 & 1.47 & 1.700 & 0.35 \\
w/o AbsPose & 41.84 & 0.72 & \best{1.04} & 1.934 & 0.51 \\
w/o BasisPrior & 43.48 & 0.96 & 1.38 & 1.638 & 0.58 \\
w/o Graph & 37.77 & 1.10 & 1.46 & 1.758 & \best{0.32} \\
w/o EndEffGeom & 38.75 & 1.14 & 1.56 & 1.862 & 0.52 \\
w/o Flow & 41.50 & 0.99 & 1.05 & 1.950 & 0.71 \\
\hline
\end{tabular}
\end{table}

\subsection{Efficiency, Qualitative, and Real-World Evaluation}

Table~\ref{tab:efficiency} reports the compact efficiency benchmark in terms of latency (Lat.), memory (Mem.), parameter footprint (Par.), and batch size (Batch). Full EAGG requires 9.68\,s of latency, 1986.44\,MB of peak GPU memory, and 52.95\,MB of model parameters at batch size 64. It is 28.2$\times$ faster than UDG and 4.4$\times$ faster than DD (pn2), while also using substantially less peak memory than those two baselines. DD (bps) is the most memory-efficient external baseline, but it has a larger 128.00\,MB model footprint and notably higher latency than EAGG.

The internal variants explain where this cost comes from. Removing the graph branch yields the lowest latency (4.89\,s) and the smallest parameter footprint (31.92\,MB), while removing the end-effector-geometry branch yields the lowest internal peak memory (1341.37\,MB). Combined with Table~\ref{tab:ablation}, these trends show that the graph and geometry branches account for a targeted computational cost in exchange for the largest quality gains.

\begin{table}[t]
\caption{Compact efficiency comparison.}
\label{tab:efficiency}
\centering
\scriptsize
\setlength{\tabcolsep}{2.5pt}
\begin{tabular}{lcccc}
\hline
\textbf{Method} & \textbf{Lat. (s)} & \textbf{Mem. (MB)} & \textbf{Par. (MB)} & \textbf{Batch} \\
\hline
UDG & 272.81 & 7322.16 & 46.85 & 16 \\
DD (pn2) & 42.43 & 2701.45 & 87.64 & 64 \\
DD (bps) & 58.90 & \best{140.53} & 128.00 & 64 \\
w/o AbsPose & 9.68 & 1986.44 & 52.95 & 64 \\
w/o LocalFeat & 7.17 & 1986.43 & 52.94 & 64 \\
w/o BasisPrior & 9.72 & 1983.54 & 49.33 & 64 \\
w/o Graph & \best{4.89} & 1898.64 & \best{31.92} & 64 \\
w/o EndEffGeom & 9.20 & 1341.37 & 32.29 & 64 \\
Full EAGG & 9.68 & 1986.44 & 52.95 & 64 \\
\hline
\end{tabular}
\end{table}

\begin{figure*}[!t]
    \centering
    \includegraphics[width=0.76\textwidth,height=0.70\textheight,keepaspectratio]{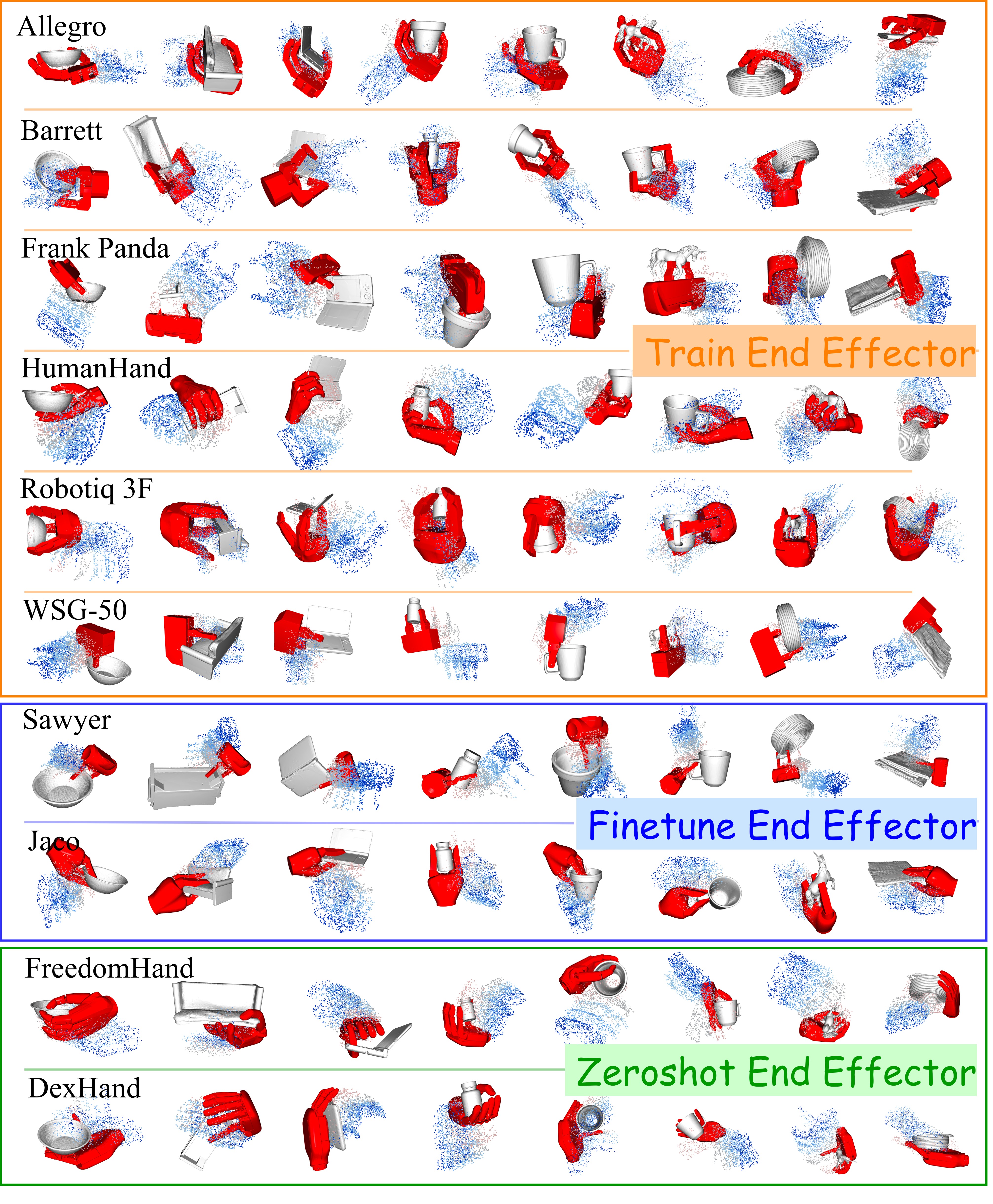}
    \caption{\textbf{Qualitative cross-end-effector grasp results.} Each row shows representative generated grasps for one end effector. The top block contains the six training end effectors used in joint base training, the middle block contains the finetuning end effectors (Sawyer and Jaco) after 10-epoch adaptation with 5\% data, and the bottom block contains the zero-shot end effectors (FreedomHand and DexHand) after lightweight adaptation seeded with SynergyGrasp \cite{niu_synergygrasp_2026} grasps on basic objects.}
    \label{fig:sim_qualitative}
\end{figure*}

Figure~\ref{fig:sim_qualitative} presents a single large qualitative montage across all 10 end effectors. The top block corresponds to the six training end effectors used in joint base training, the middle block corresponds to the two finetuning end effectors after 10-epoch adaptation, and the bottom block corresponds to the two zero-shot end effectors after lightweight adaptation seeded with SynergyGrasp grasps \cite{niu_synergygrasp_2026}. The figure shows that the unified model does not collapse to a single closure template. Training end effectors produce visibly different strategies on the same object class, the finetuning end effectors recover coherent multi-contact configurations after limited supervision, and the zero-shot end effectors already exhibit plausible grasp families after lightweight adaptation.

The remaining weak cases are structured rather than arbitrary. Large objects continue to challenge Franka Panda and WSG-50 because nearly one-dimensional closing motion leaves little room to correct pose error once the wrist approach is fixed. The hardest zero-shot cases instead arise from late-stage finger ordering conflicts or partial enclosure on geometry that is underrepresented in the lightweight adaptation set. These qualitative patterns are consistent with the quantitative regime ordering in Tables~\ref{tab:main_results}--\ref{tab:ablation} and with the geometry analysis in Fig.~\ref{fig:igi_diagnostic}.

Table~\ref{tab:real_world} reports hardware trials on three platforms across five object groups. UR5 + FreedomHand completes 62 trials with 91.94\% average success, UR5 + DaHuan AG95 completes 60 trials with 95.00\% average success, and SOARM101 completes 64 trials with 89.06\% average success. FreedomHand remains above 84.6\% on every object group, DaHuan AG95 reaches 100\% on groups A and B, and SOARM101 shows the largest cross-group variation.

\begin{figure*}[t]
    \centering
    \includegraphics[width=0.65\linewidth]{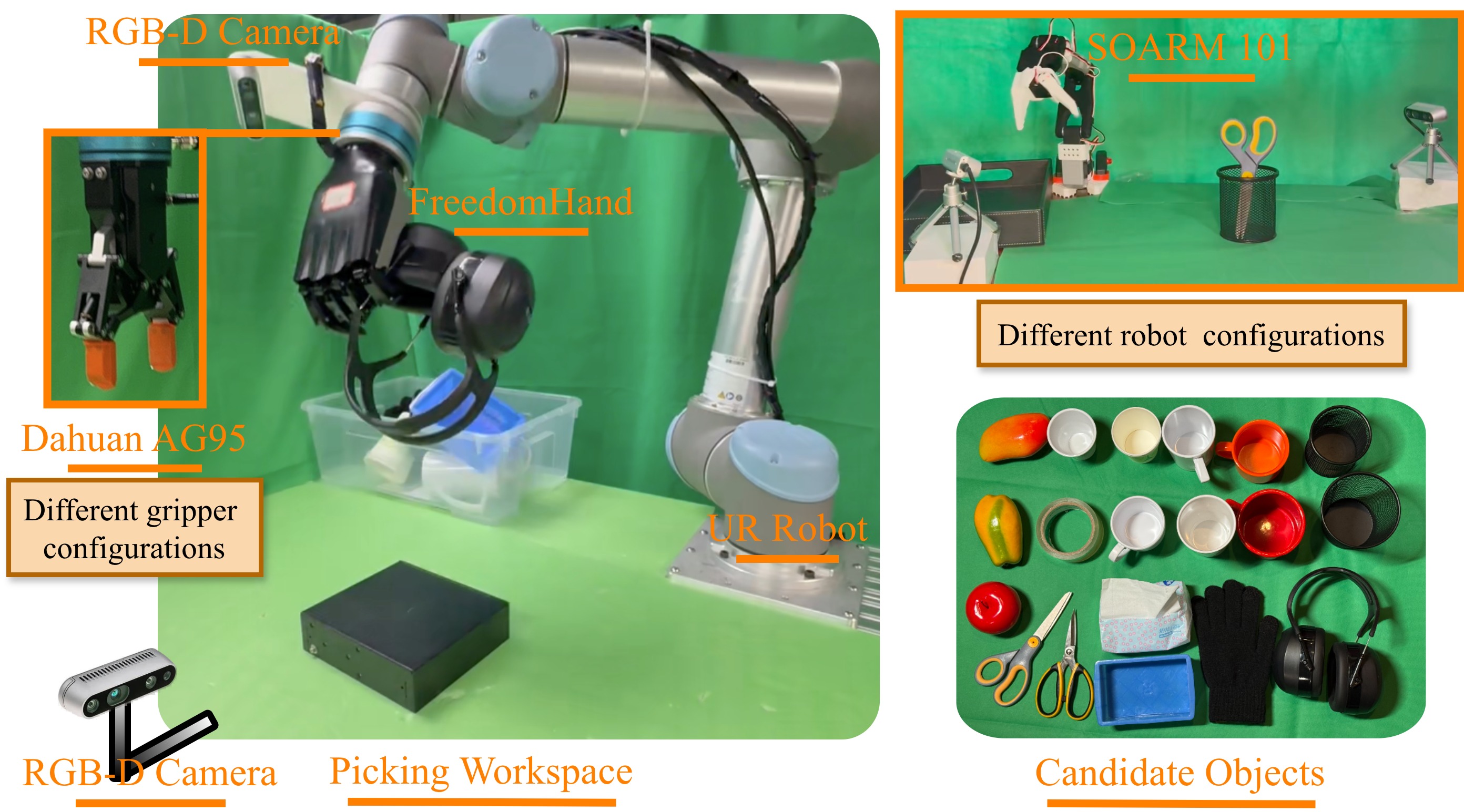}
    \caption{\textbf{Real-world hardware configurations.} The hardware evaluation uses a UR-based workstation with interchangeable FreedomHand and DaHuan AG95 end effectors, an SOARM101 platform, RGB-D sensing, and the representative object set shown in the figure.}
    \label{fig:hardware}
\end{figure*}

\begin{figure*}[t]
    \centering
    \includegraphics[width=0.92\textwidth]{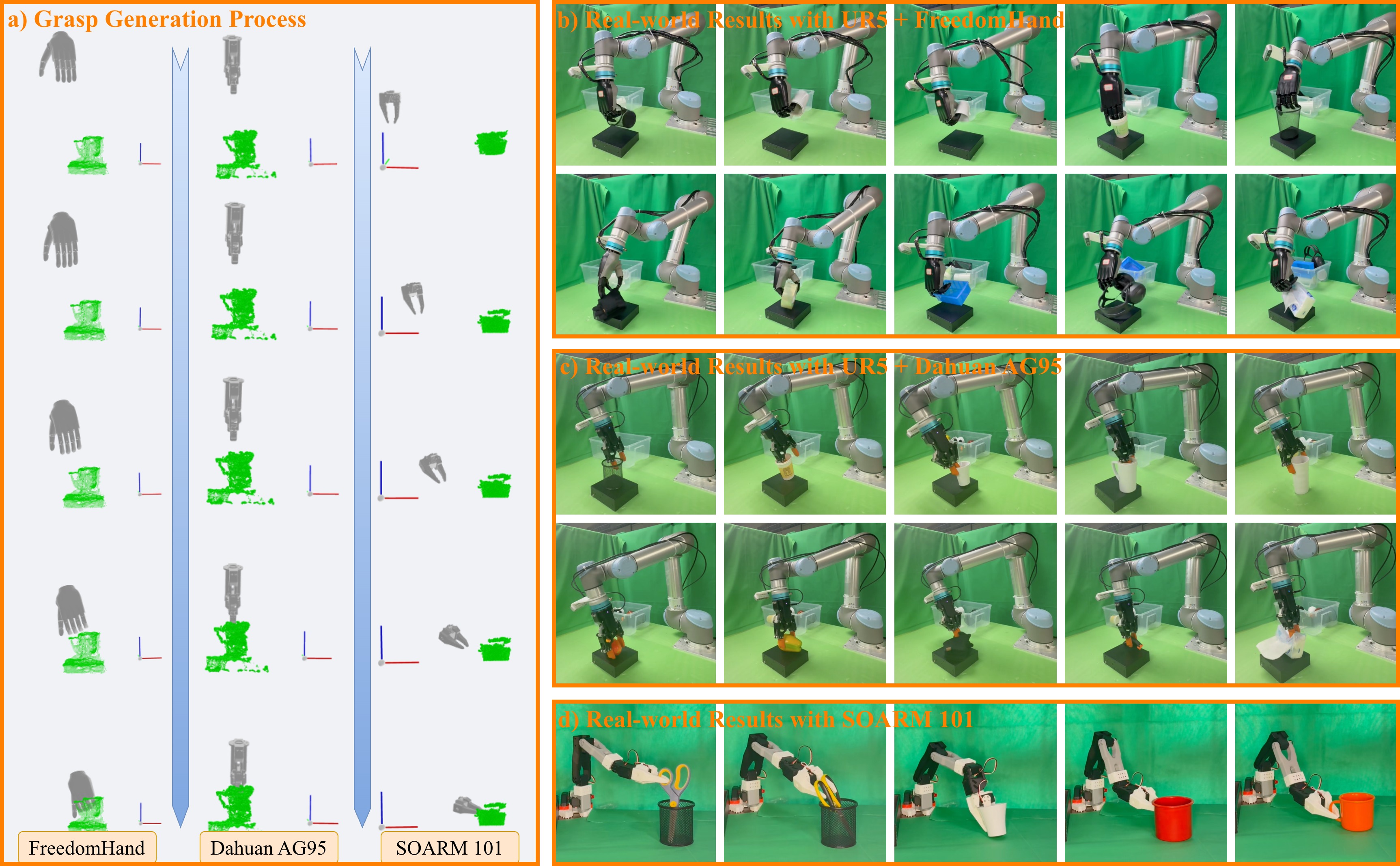}
    \caption{\textbf{Representative real-world execution results.} Panel (a) shows generated grasps for FreedomHand, DaHuan AG95, and SOARM101. Panels (b)--(d) show representative successful executions on UR5 + FreedomHand, UR5 + DaHuan AG95, and SOARM101, respectively.}
    \label{fig:real_gallery}
\end{figure*}

Figure~\ref{fig:hardware} summarizes the hardware configurations, and Fig.~\ref{fig:real_gallery} shows representative generated grasps and successful executions. Together with Table~\ref{tab:real_world}, these results show that the learned policy transfers across robot arms, end effectors, and sensing setups without redesigning the generator for each platform.

The physical results sharpen the meaning of the simulation benchmark. The shared representation produces grasps that remain executable under real sensing noise, robot-controller delay, and embodiment-specific calibration error, which is precisely the operating regime in which cross-end-effector grasp generation is most valuable.

This hardware evidence also complements the simulation metrics. Simulation reveals how embodiment alignment affects contact distance, penetration, and adaptation behavior, whereas hardware trials test whether those improvements survive perception noise and actuation uncertainty.

\begin{table*}[t]
\caption{Real-world evaluation on three hardware platforms across five object groups.}
\label{tab:real_world}
\centering
\footnotesize
\setlength{\tabcolsep}{4.2pt}
\begin{tabular}{llccc}
\hline
\textbf{Setup} & \textbf{Group} & \textbf{Objects} & \textbf{Attempts} & \textbf{SR (\%)} \\
\hline
\multirow{6}{*}{UR5 + FreedomHand}
& A & 10 & 11 & 90.91 \\
& B & 11 & 11 & \best{100.00} \\
& C & 13 & 14 & 92.86 \\
& D & 12 & 13 & 92.31 \\
& E & 11 & 13 & 84.62 \\
& Average & 57 & 62 & 91.94 \\
\hline
\multirow{6}{*}{UR5 + DaHuan AG95}
& A & 10 & 10 & \best{100.00} \\
& B & 11 & 11 & \best{100.00} \\
& C & 13 & 14 & 92.86 \\
& D & 12 & 13 & 92.31 \\
& E & 11 & 12 & 91.67 \\
& Average & 57 & 60 & \best{95.00} \\
\hline
\multirow{6}{*}{SOARM101}
& A & 10 & 11 & 90.91 \\
& B & 11 & 12 & 91.67 \\
& C & 13 & 15 & 86.67 \\
& D & 12 & 14 & 85.71 \\
& E & 11 & 12 & 91.67 \\
& Average & 57 & 64 & 89.06 \\
\hline
\end{tabular}
\end{table*}

\section{Conclusion}

This paper presented EAGG for cross-end-effector grasp synthesis. EAGG works by aligning three forms of embodiment structure inside one generator: an end-effector-specific low-dimensional control basis, a topology-aware graph that preserves embodiment organization, and geometry-aware conditioning refreshed during sampling through IGI. This combination allows one model to operate across heterogeneous end effectors without flattening them into a shared raw joint parameterization.

The experimental results support a consistent interpretation. EAGG remains within 1.10 SR points of specialized training on the six training end effectors while preserving transfer to finetuning and zero-shot embodiments, and the ablations show that topology-aware conditioning and dynamic geometry are the dominant factors behind that performance. The central insight is not to remove embodiment structure, but to align it with object geometry throughout generation.

The current evaluation is limited to a fixed set of end effectors and objects, and IGI yields smaller gains on low-DoF grippers whose geometry changes little during closure. Even with these boundaries, EAGG shows that embodiment alignment is a practical route to unified grasp generation across heterogeneous robotic end effectors.

The results also suggest a useful design principle for multi-embodiment manipulation systems. When heterogeneous end effectors must share one generator, a promising abstraction is a shared model whose internal representation preserves embodiment differences in an aligned form. This perspective complements more morphology-agnostic approaches by offering an alternative when embodiment-specific structure remains important. Under this view, transfer improves because the model learns which aspects of grasping should be shared and which should remain embodiment specific. In this sense, EAGG is relevant not only as a grasp generator, but also as one practical template for broader multi-embodiment generative policies.


\end{document}